\documentclass[draftclsnofoot,onecolumn]{IEEEtran}
\IEEEoverridecommandlockouts
\pdfoutput=1

\usepackage{cite}
\usepackage{amsmath,amssymb,amsfonts}

\usepackage{algorithmic}
\usepackage{graphicx}
\usepackage{subfigure}
\usepackage[justification = centering,labelsep = period]{caption}
\usepackage{}
\usepackage{mathtools}
\usepackage{textcomp}
\usepackage{xcolor}
\usepackage{booktabs}
\newcommand{\DOI}[1]{doi: #1}
\def\BibTeX{{\rm B\kern-.05em{\sc i\kern-.025em b}\kern-.08em
    T\kern-.1667em\lower.7ex\hbox{E}\kern-.125emX}}
\begin{document}

\title{A Generative Graph Contrastive Learning Model with Global Signal\\}

\author{\IEEEauthorblockN{Xiaofan Wei\IEEEauthorrefmark{1}, Binyan Zhang\IEEEauthorrefmark{2}} \\
\IEEEauthorblockA{\IEEEauthorrefmark{1} \textit{College of Computer and Information Science, Southwest University, Chongqing, China}\\
 \IEEEauthorblockA{\IEEEauthorrefmark{2} \textit{College of Han Hong, Southwest University, Chongqing, China}}}}
\maketitle

\begin{abstract}
Graph contrastive learning (GCL) has garnered significant attention recently since it learns complex structural information from graphs through self-supervised learning manner. However, prevalent GCL models may suffer from performance degradation due to inappropriate contrastive signals. Concretely, they commonly generate augmented views based on random perturbation, which leads to biased essential structures due to the introduction of noise. In addition, they assign equal weight to both hard and easy sample pairs, thereby ignoring the difference in importance of the sample pairs. To address these issues, this study proposes a novel Contrastive Signal Generative Framework for Accurate Graph Learning (CSG\textsuperscript{2}L) with the following two-fold ideas: a) building a singular value decomposition (SVD)-directed augmented module (SVD-aug) to obtain the global interactions as well as avoiding the random noise perturbation; b) designing a local-global dependency learning module (LGDL) with an adaptive reweighting strategy which can differentiate the effects of hard and easy sample pairs. Extensive experiments on benchmark datasets demonstrate that the proposed CSG\textsuperscript{2}L outperforms the state-of-art baselines. Moreover, CSG\textsuperscript{2}L is compatible with a variety of GNNs.
\end{abstract}


\section{Introduction}
With the development of information technology, graph data has been widely used in  bioinformatics networks\cite{b43,b57,b58}, recommendation systems\cite{b55,b64,b65,b66,b67,b68,b69}, and transportation networks\cite{b59,b60,b61,b62,b63}. Graph data are usually represented by a complex association structure of nodes and edges, where nodes represent entities and edges describe the relationships between entities\cite{b70,b71,b72,b73,b74,b75,b76}. In this context, how to effectively learn the representation of graph data has become a core challenge\cite{b44,b45,b46,b47,b48,b49,b50,b51,b52,b53,b54}.

Graph neural networks (GNNs) have shown great performance in recent years due to their ability to effectively utilize graph structure and node feature information\cite{b32,b33,b34,b35,b36}. They have been widely applied in tasks such as node classification \cite{b26,b27,b28,b29,b30,b31,b56}, link prediction \cite{b1,b83,b84}, and graph classification \cite{b2}. The core idea of GNN is to aggregate information within the neighborhood of a graph through a message passing mechanism, thereby learning a representation that can capture local graph structure and node features \cite{b5,b6,b7,b85}. This mechanism allows the GNN model to simultaneously utilize node feature information and the topological structure of the graph when performing tasks\cite{b37,b38,b39,b40,b41,b42}. However, existing GNN methods face several limitations. For example, in sparse graphs, neighborhood information is often insufficient to provide meaningful signals for representation learning, and traditional GNNs only focus primarily on the local neighborhoods, this local dependency characteristic limits the model’s learning capabilities and therefore makes it difficult to learn robust node representations. In addition, since graph data usually comes from real-world application scenarios\cite{b77,b78,b79,b80,b81,b82}, it often contains noisy edges or abnormal nodes. In this case, the simple local feature aggregation mechanism cannot effectively distinguish useful information from noise, resulting in inaccurate graph learning by the model, affecting the final task effect.

To address these issues, GCL has received increasing attention as an emerging self-supervised learning method \cite{b14,b15,b16,b17,b18}. The core idea of GCL is to construct positive and negative sample pairs and learn the representation of nodes or graphs by pulling positive pairs closer and pushing negative pairs apart, thereby improving the quality of representation learning \cite{b15}. Unlike traditional supervised learning methods, GCL does not rely on manually labeled data and can mine useful information from unlabeled data \cite{b3}. In addition, GCL can alleviate the problems of data sparsity and missing labels to a certain extent by comparing positive and negative sample pairs \cite{b4}.

Some studies have shown that contrastive learning can benefit from hard sample pairs that are indistinguishable from anchor samples \cite{b19}. Distinguishing between hard and easy sample pairs is crucial for enhancing the effectiveness of contrastive learning. Hard sample pairs are those that are misclassified during training and contain valuable information that can help the model capture complex relationships within the graph. Learning from these pairs allows the model to improve its ability to discriminate between subtle differences. For easy sample pairs that the model can easily and correctly classify, their learning value is relatively low because they have limited contribution to the discrimination task.

Based on some existing research, we found that they usually face two main problems:

1. When generating contrastive views, they often rely on random perturbations or simple graph structure transformations. Although this strategy can generate diverse contrastive views, it may lose some important graph structure information and introduce noise, which hinder the model’s ability to learn accurate contrastive signals.

2. During the training process, all hard and easy sample pairs are usually assigned the same weight, assuming that all sample pairs contribute equally to the learning of the model, ignoring the difference between hard sample pairs and easy sample pairs. Hard sample pairs contain more useful information, and simply treating all sample pairs equally often leads to the model failing to fully learn the complex structure and potential information of hard samples, limiting the effect of graph learning.

To overcome these challenges, we propose CSG\textsuperscript{2}L, a novel graph contrastive signal generative framework which consists of two essential components: a SVD-directed augmented module (SVD-aug) and a local-global dependency learning module (LGDL). The SVD-aug module captures global collaborative signals, while the LGDL module integrates local and global information. Specifically, in the SVD-aug module, we first perform SVD on the original graph structure to obtain a low-rank approximation of the graph, thereby generating an augmented graph containing global information. On this basis, the LGDL module constructs contrastive sample pairs by using the augmented graph and the original graph. This approach ensures that the model learns both local and global graph interactions. To further optimize the learning process, we introduce an adaptive reweighting strategy into the LGDL module that assigns greater importance to hard sample pairs during training. Through these methods, the model can generate more accurate graph contrastive signals and improve the robustness of learned representations.

Our contributions are as follows:

\begin{itemize}
\item We propose a novel graph contrastive signal generative framework CSG\textsuperscript{2}L which can be combined with existing GNN models. It includes two modules, i.e., SVD-aug and LGDL, to effectively generate accurate graph contrastive signals.
\item We build a SVD-directed augmented module (SVD-aug). By utilizing SVD-directed global augmentation, CSG\textsuperscript{2}L alleviates the impact of noisy contrastive signals and captures both local and global interactions.
\item We design a local-global dependency learning module (LGDL) by introducing an adaptive reweighting strategy, which can effectively improve the performance of CSG\textsuperscript{2}L in the contrastive learning process and generate more accurate contrastive signals.
\item We conduct extensive experiments on six real datasets, and the results show that CSG\textsuperscript{2}L significantly improves the performance of GNN models, fully proving the superiority of our framework.
\end{itemize}

\section{Related Work}

\subsection{Graph Neural Networks}\label{AA}
Graph neural networks have made significant progress in processing non-Euclidean structured data, with wide applications in node classification, link prediction, and graph classification. Classic models include GCN \cite{b5}, GraphSAGE \cite{b6} and GAT \cite{b7} etc., which capture local structural through efficient neighbor aggregation mechanisms. GCN uses the spectral method to perform weighted averaging of neighbor node features, aiming to improve the expressiveness of node representation through Laplace smoothing. GraphSAGE introduces a neighbor sampling method on this basis, thereby improving the scalability to large-scale graph. GAT introduces the attention mechanism, allowing the model to assign different weights to different neighbor nodes, and then dynamically adjusting the effect of message passing. Although these methods perform well on many tasks, they primarily rely on local information aggregation, which can lead to the neglect of global structural patterns in the graph, particularly obvious in scenarios with sparse data or high noise. To address this limitation, recent studies have explored enhancing GNNs with global graph signals. For example, The SIGN \cite{b8} model injects global information into the node representation through a global feature expansion mechanism, effectively capturing low-frequency signals; GloGNN \cite{b9} uses spectral methods to extract low-frequency structural information of the graph, enhancing the model’s ability to capture global collaborative relationship. In addition, ChebNet \cite{b10} approximates the Laplace matrix through Chebyshev polynomials, thereby capturing global structural information while retaining local information; SGC \cite{b11} further simplifies GCN and directly aggregates global neighbor node information through linearized convolution operations, effectively reducing computational complexity. In addition to spectral domain methods, some researchers have also explored the effective use of global information in the spatial domain. For example, Graph Heat Kernel \cite{b12} and PPR (Personalized PageRank) \cite{b13} define a global propagation matrix based on the propagation mechanism between nodes to capture the collaborative relationship between long-distance nodes. These works have made progress in global structure modeling, but how to efficiently combine global information with local information remains an issue worthy of further exploration.

\subsection{Graph Contrastive Learning}
\begin{figure*}[t!]
\centerline{\includegraphics[width=16cm]{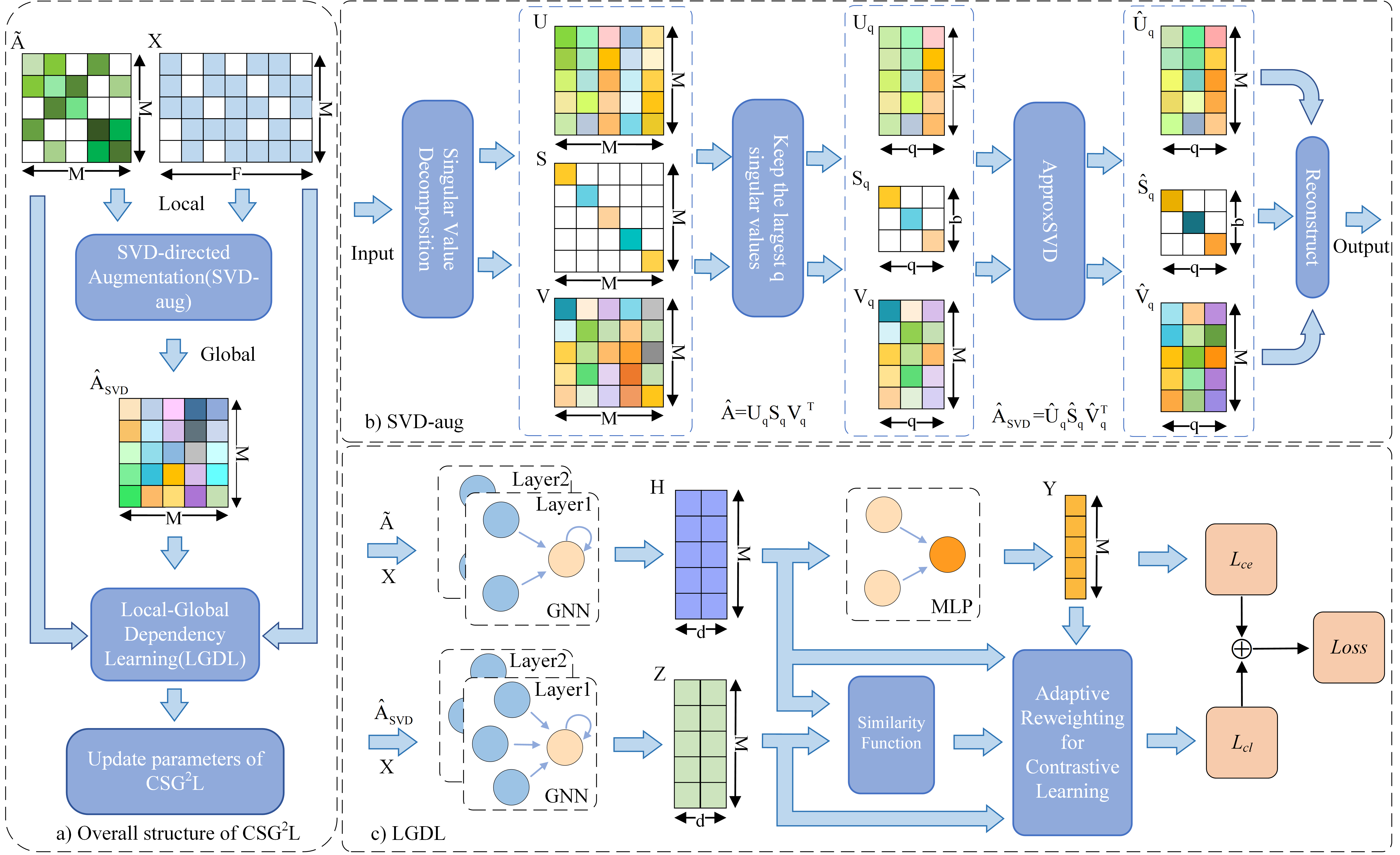}}
\captionsetup{justification=raggedright,singlelinecheck=false,font=footnotesize}
\caption{a) The overall structure of CS$\mathrm{G}^2$L. It includes two components: b) SVD-aug module performs SVD-directed reconstruction of the original graph to obtain augmented graph. c) LGDL module inputs both graphs into a shared GNN, and further input the embedding of original graph into an MLP classifier to compute the classification loss. In addition, an adaptive reweighting strategy is introduced for contrastive learning. Finally, the overall objective function can be obtained.}
\label{fig1}
\end{figure*}
As a self-supervised learning method, graph contrastive learning can effectively mine the latent features of graphs and reduce dependence on manual labels. Existing research mainly focuses on the generation of contrastive views and the optimization of contrastive loss functions, and proposes a variety of innovative solutions to improve model performance. In terms of view generation, GraphCL \cite{b14} generates different contrasting views by randomly perturbing nodes, edges, and features, enhancing the diversity of the model. GCA \cite{b15} further introduces an enhancement strategy based on importance weights to avoid the risk of invalid perturbations. However, these methods generally rely on randomness and not preserve the global interactions of the graph effectively. In addition, these augmented methods are sensitive to noise, and their performance will decrease significantly especially with noisy or low-quality data, therefore generate inaccurate contrastive signals. For contrastive loss, InfoNCE loss is one of the most widely used loss function. It effectively improves the distinguishability of node representation by maximizing the similarity of positive sample pairs and minimizing the similarity of negative sample pairs \cite{b14,b15}, but this method usually gives the same weight to all sample pairs, ignoring the importance of hard sample pairs for model learning. In order to improve the learning ability of hard sample pairs, SUGAR \cite{b16} effectively improves the quality of contrastive learning signals by comparing the potential distribution information of the graph. BGRL \cite{b17} simplifies the optimization process by removing negative sample pairs. ProGCL \cite{b18} introduce a refined approach to build a more suitable measure for the hard negative sample pairs together with similarity. Although these methods have achieved certain results, the weight allocation of hard and easy sample pairs still lacks flexibility, and it is difficult to effectively capture the importance of hard sample pairs, leading to inaccurate graph learning.

\section{Methodology}
We now introduce our CSG\textsuperscript{2}L model, which captures global interactions and introduces an adaptive reweighting strategy into contrastive learning, aiming to address the main challenges associated with contrastive learning for accurate graph learning. As shown in Fig.~\ref{fig1}, CSG\textsuperscript{2}L consists of two components: a SVD-directed augmented module (SVD-aug) and a local-global dependency learning module (LGDL). In this section, we first define some basic concepts and then discuss the details of the two components.

\subsection{Problem Defination}
We consider a graph $G=(N,E,X,A)$, where $N=\{n_1,n_2,\cdots,n_M\}$ and $E\subseteq N\times N$ represent the set of nodes and the set of edges, $M$ is the number of nodes. The node feature matrix is denoted as $X \in \mathbb{R}^{M\times F}$, where $F$ is the number of node feature dimensions. The adjacency matrix is denoted as $A=a_{ij} \in \{0,1\}^{M\times M}$, $a_{ij}=1$ when $(n_i,n_j)\in N$, the diagonal matrix is $D=diag(d_1,d_2,\cdots,d_M)\in \mathbb{R}^{M\times M}$. And we define the normalized  adjacency matrix as $\Tilde{A}\in \mathbb{R}^{M\times M}$ and $\Tilde{A}=D^{-1/2}(A+I)D^{-1/2}$, where $I\in \mathbb{R}^{M\times M}$ is an identity matrix.

\subsection{SVD-directed Augmented Module}

Global interactions plays a crucial role in capturing long-range dependencies and enhancing the expressiveness of node representations in complex graph structures. By effectively utilizing global interactions, models can better understand long-distance relationships between nodes, leading to improved performance in graph learning tasks. However, existing graph contrastive learning methods usually rely on random augmentations with local neighborhood information, ignoring the global structure of the graph. To address these limitations, we build a SVD-directed augmented module (SVD-aug) to extract global information from the original graph and generate an augmented view that incorporate richer global interactions. Specifically, SVD-aug consists of two steps: graph reconstruction and approximation of SVD.

\textit{\textbf{Graph Reconstruction}}. To effectively capture the global interactions and enhance the ability to model global signals, we adopts SVD \cite{b20} to extract important graph structure information from the global perspective. Specifically, we perform SVD on the normalized adjacency matrix $\Tilde{A}$, decomposing it into two orthogonal matrices and a singular value matrix, denoted as $\Tilde{A}=USV^T$. Here, $U$ and $V$ are both $M\times M$ standard orthogonal matrices, with their columns representing the eigenvectors of the row-row and column-column correlation matrix of $\Tilde{A}$. $S$ is an $M\times M$ diagonal matrix storing the singular values of $\Tilde{A}$, which reflect the important information of the matrix in different directions. The singular vectors corresponding to the top-ranked singular values usually contain the main information. The largest singular values are associated with the principal components of the matrix. Retaining the largest singular values and their corresponding singular vectors preserve the key information of the graph while reducing noise and ensuring effective reconstruction. Thus, we truncate the singular value list to keep the largest q singular values. Using these truncated singular values and their corresponding singular vectors, we reconstruct the normalized adjacency matrix as $\hat{A}=U_qS_qV_q^T$, where $U_q\in \mathbb{R}^{M\times q}$ and $V_q\in \mathbb{R}^{M\times q}$ contain the first q columns of $U$ and $V$ respectively, $S_q\in \mathbb{R}^{q\times q}$ is the diagonal matrix of the q largest singular values. The reconstructed matrix $\hat{A}$ is a low-rank approximation of the normalized adjacency matrix $\Tilde{A}$.

\textit{\textbf{Approximation of SVD}}. However, performing exact SVD on the normalized adjacency matrix of large graphs is highly expensive and may result in significant loss of efficiency. Therefore, we use the randomized algorithm which is usually used in the ApproxSVD algorithm \cite{b21}. This method first transform the original matrix into a low-rank orthogonal matrix, and continuously approximate the range of the original matrix, and then SVD is applied to this smaller-scale matrix:
\begin{equation}
\hat{U}_q,\hat{S}_q,\hat{V}_q^T=ApproxSVD(\Tilde{A},q),\   \hat{A}_{SVD}=\hat{U}_q\hat{S}_q\hat{V}_q^T.\label{eq1}
\end{equation}

The graph structure learning based on SVD can effectively reduce the influence of noise in the network, demonstrating strong robustness in most scenarios, and can also maintain the global information while considering the characteristics of each sample pair.

\subsection{Local-Global Dependency Learning}\label{AAA}
In graph contrastive learning, accurate contrastive signals are crucial for model learning. Effective contrastive signals can not only improve model performance but also help discriminate subtle differences between sample pairs more effectively. The combination of local and global information plays a critical role in generating high-quality contrastive signals. Local information captures neighborhood relationship, while global information provides an overall perspective of the graph. The combination of the two can improve the model’s discriminative ability, leading to more effective graph representation learning. However, most existing graph contrastive learning methods usually focus on generating randomly augmented views and then treat all sample pairs with equal importance during training. This approach does not consider local and global information which may result in inaccurate contrastive signals and hinder model performance. To address this issue, we design a local-global dependency learning module (LGDL) to integrate local and global information, improve the discrimination between hard and easy sample pairs, and then generate more accurate contrastive signals. Specifically, LGDL contains two steps: local-global encoding and adaptive reweighting contrastive learning.

\textit{\textbf{Local-Global Encoding}}. GNNs capture the complex structural features in graphs by propagating information between neighbors and integrating local structure and features into high-dimensional node embeddings. In this module, we calculate node embeddings based on the original graph and the reconstructed graph respectively to better combine global information and local relationship between nodes, providing stronger representation capabilities for contrastive learning. Specifically, we input the original graph node features $X$ and normalized adjacency matrix $\Tilde{A}$ into a GNN model $f(\cdot,\cdot)$, and calculate the node embeddings through two-layer message propagation:
\begin{equation}
H=f(\Tilde{A},X),\label{eq2}
\end{equation}
where $H\in \mathbb{R}^{M\times m}$ represents node embeddings in the original graph, and $m$ is the dimension of the embedding. In our framework, a variety of GNN models can be used, such as GCN \cite{b5}, GIN \cite{b25}, and GPRGNN \cite{b26}. Then, we rewrite the message propagation rule in equation \eqref{eq2} with the approximated matrices in equation \eqref{eq1}:
\begin{equation}
Z=f(\hat{A}_{SVD},X)=f(\hat{U}_q\hat{S}_q\hat{V}_q^T,X),\label{eq3}
\end{equation}
where $Z\in \mathbb{R}^{M\times m}$ is node embeddings of the reconstructed graph. In this way, the nodes high-order features can further integrate local and global information during the propagation process of the GNN.

\textit{\textbf{Adaptive Reweighting Contrastive Learning}}. Graph contrastive learning is a self-supervised learning method that can improve the model's ability to discriminate features by measuring the similarity of sample pairs. In traditional graph contrastive learning, InfoNCE loss is a commonly used optimization objective. Its key idea is to learn effective representations by minimizing the distance between positive sample pairs and maximizing the distance between negative ones. The InfoNCE loss is formulated as:
\begin{equation}\begin{aligned}
l_{info}(h_i,z_i)&=\\&-\log\frac{e^{\theta(h_i,z_i)/\tau}}{e^{\theta(h_i,z_i)/\tau}+\sum\limits_{k\ne i}(e^{\theta(h_i,z_k)/\tau}+e^{\theta(h_i,h_k)/\tau})},\label{eq4}
\end{aligned}
\end{equation}
where $h_i$,$z_i$ denote the embeddings of the $i$-th sample on the original graph and the reconstructed graph, and $\tau$ is a temperature parameter. Besides, we define $\theta(h_i,z_i)=s(g(h),g(z))$, $s(\cdot,\cdot)$ is the cosine similarity and $g(\cdot)$ is a non-linear projection to enhance the expression power of the critic function. The disadvantage of InfoNCE loss is that it treats both hard sample pairs and easy sample pairs equally. This strategy may lead to insufficient learning for hard samples which can weaken the model’s representation learning. At the same time, over-optimization of easy samples wastes computing resources and limits the model’s discriminative capability. To address these issues, we introduce a reweighting function R to adaptively adjust the weights of sample pairs. Concretely, after encoding, the node embeddings H learned from the original graph are input into an MLP classifier $c(\cdot)$ to obtain the prediction $P=c(H)\in \mathbb{R}^{M\times C}$, where $C$ denotes the number of classes. Since the prediction may be noisy, we only consider the nodes with the most confident prediction for each class, that is, for the high-confidence nodes whose prediction values for a specific class exceed a given threshold, we calculate their pseudo-labels $Y\in \mathbb{R}^M$ based on $P$, and then use $Y$ to obtain the sample pair pseudo-label matrix $Q\in \mathbb{R}^{M\times M}$ as follows:
\begin{equation}
Q_{ik}=\begin{cases} 1, & Y_i=Y_k, \\ 0, & Y_i\ne Y_k,\end{cases}\label{eq5}
\end{equation}
where $Q_{ik}$ represents the pseudo-label relationship between the $i$-th sample and the $k$-th sample, that is, when the pseudo-labels are the same ($Q_{ik}=1$), they are more likely to be the positive sample pair, and when the pseudo-labels are not the same ($Q_{ik}=0$), it means that they are more likely to be the negative sample pair.

Based on $s$ and the generated pseudo-label matrix $Q$, we define the reweighting function $R$ as:
\begin{equation}
R(h_i,z_k)=\begin{cases} 1, \ \ others, \\ \left|Q_{ik}-Norm(s(h_i,z_k))\right|, & i,k\in O,\end{cases}\label{eq6}
\end{equation}
where $Norm(\cdot)$ denotes the min-max normalization and $O$ denotes the high-confidence nodes. When the sample pair is not with high confidence, we keep the original InfoNCE setting for it. For high-confidence sample pairs, i.e., $i,k\in O$, their weights are adaptively reweighted with the pseudo-labels and the samples similarity. Concretely, when the $i$-th and $k$-th sample nodes are identified as a positive sample pair ($Q_{ik}=1$), if the calculated similarity of their embeddings after passing through the GNN model is low, it means that the positive sample pair is likely to be a hard sample pair. Therefore, a larger weight is given to the sample pair, and the positive sample pairs with high similarity (the easy sample pairs) are down-weighted. For negative sample pair with different pseudo-labels ($Q_{ik}=0$), if their embedding similarity is low, it means that the negative sample pair is likely to be an easy sample pair, so a smaller weight is given to the sample pair, while negative sample pairs with higher similarity (the hard sample pairs) are up-weighted. This adaptive reweighting strategy focuses more on hard sample pairs, less on easy sample pairs, thereby further improving the discriminative capability of sample pairs and generating more accurate contrastive signals.

Based on $R$ and $\theta$, the reweighted InfoNCE can be defined as follows:
\begin{equation}\begin{aligned}
&l(h_i,z_i)=-\log(\\&\frac{e^{(R(h_i,z_i)\cdot\theta(h_i,z_i))/\tau}}{\splitfrac{e^{(R(h_i,z_i)\cdot\theta(h_i,z_i))/\tau}}{+\sum\limits_{k\ne i}e^{(R(h_i,z_k)\cdot\theta(h_i,z_k))/\tau}+\sum\limits_{k\ne i}e^{(R(h_i,h_k)\cdot\theta(h_i,h_k))/\tau}}}),\label{eq7}
\end{aligned}
\end{equation}

The overall contrastive learning objective is the average over all positive pairs, formally given by:
\begin{equation}
L_{cl}=\frac{1}{2M}\sum_{i=1}^M\left[l(h_i,z_i)+l(z_i,h_i)\right].\label{eq8}
\end{equation}

Finally, we jointly optimize the main objective function of the node classification model with the contrastive loss:
\begin{equation}
L=L_{ce}+\lambda L_{cl},\label{eq9}
\end{equation}
where $L_{ce}$ denotes the cross entropy loss and $\lambda$ is the weight of the contrastive loss
$L_{cl}$.

\begin{table}[t]
\captionsetup{labelsep=newline, font=footnotesize}
\caption{\textsc{Datasets statistics}}
\begin{center}
\begin{tabular}{l c c c c}
\toprule
\textbf{Datasets}&\textbf{\#Nodes}&\textbf{\#Edges}&\textbf{\#Features}&\textbf{\#Classes} \\
\midrule
Wisconsin & 251 & 499 & 1703 & 5 \\
Texas & 183 & 309 & 1703 & 5 \\
Cornell & 183 & 295 & 1703 & 5 \\
Chameleon & 2277 & 36101 & 2325 & 5 \\
Cora & 2708 & 5429 & 1433 & 7 \\
Citeseer & 3327 & 4732 & 3703 & 6 \\
\bottomrule
\end{tabular}
\label{tab1}
\end{center}
\end{table}

\section{Experiments}
In this section, we evaluate the performance of the CSG\textsuperscript{2}L framework on node classification tasks on six real-world benchmark datasets. We first introduce the datasets, baseline models, and hyper-parameter settings. Then, we compare CSG\textsuperscript{2}L with the baseline models and analyze the results. Finally, hyper-parameter analysis and comprehensive ablation study are performed to thoroughly demonstrate the effectiveness of our method.

\subsection{Datasets}
We evaluate the proposed CSG\textsuperscript{2}L on six widely used real-world datasets, including four heterophilic graph datasets Wisconsin, Texas, Cornell \cite{b22}, and Chameleon \cite{b23}, and two homophilic graph datasets Cora and Citeseer \cite{b24}. Table~\ref{tab1} summarizes the characteristics of all datasets in detail. For all benchmarks, we use the same feature vectors, graph structures, and class labels provided in \cite{b22}, and perform standard 10 fixed random split of the datasets (60\%/20\%/20\% of nodes per class for training/validation/testing). When the validation accuracy of the trained model reaches a maximum value of each run, we start the testing procedure and calculate the average test accuracy of these ten runs.

\subsection{Baselines}
\begin{table*}[t!]
\captionsetup{labelsep=newline, font=footnotesize}
\caption{\textsc{Average accuracy(\%) and standard deviation for ten different data splits on six real-world graph datasets. The best results are highlighted in bold. The improvement of the CSG\textsuperscript{2}L model relative to its counterpart is calculated, with the improvement is indicated by an upward arrow ↑.}}
\begin{center}
\begin{tabular}{l c c c c c c c}
\toprule
\textbf{Method}&\textbf{Wisconsin}&\textbf{Texas}&\textbf{Cornell}&\textbf{Chameleon}&\textbf{Cora}&\textbf{Citeseer}&\textbf{Average} \\
\midrule
GAT & 56.31±0.55 & 58.27±1.28 & 56.59±2.41 & 54.69±1.95 & 86.13±0.84 & 75.46±1.37 & 64.58 \\
GraphSAGE & 79.09±3.23 & 78.02±4.71 & 67.12±4.27 & 58.73±1.68 & 86.30±1.02 & 76.14±0.98 & 74.23 \\
MixHop & 75.88±4.90 & 77.84±7.73 & 73.51±6.34 & 60.50±2.53 & 83.10±2.03	& 73.05±2.65 & 73.98 \\
H2GCN & 82.08±3.22 & 79.70±5.16 & 78.16±4.05 & 56.85±1.68 & 86.26±1.08 & 75.51±1.33 & 76.43 \\
GOAL & 73.55±3.04 & 74.77±4.11 & 73.88±3.29 & 67.32±0.66 & 85.92±0.80 & 74.77±0.46 & 75.04 \\
OrderedGNN & 79.46±3.79 & 78.31±4.64 & 79.47±1.18 & 65.28±1.15 & 84.48±0.91 & 73.38±0.67 & 76.73 \\
ES-GNN & 75.50±4.21 & 75.47±2.15 & 74.06±4.57 & 67.58±0.63 & 84.17±1.12 & 73.47±0.56 & 75.04 \\
\midrule
GPRGNN & 83.25±0.63 & 81.31±1.03 & 79.93±0.92 & 67.07±0.53 & 85.32±0.26 & 76.06±0.53 & 78.82 \\
\textbf{CS$\mathrm{\textbf{G}}^2$$\textbf{L}_{\textit{GPRGNN}}$} & \textbf{86.08±1.53} & \textbf{85.57±0.66} & \textbf{82.98±0.64} & \textbf{68.12±1.02} & \textbf{86.89±0.49} & \textbf{76.79±0.27} & \textbf{81.07} \\
\textit{improvements} & ↑2.83 & ↑4.26 & ↑3.05 & ↑1.05 & ↑1.57 & ↑0.73 & ↑2.25 \\
\bottomrule

\end{tabular}
\label{tab2}
\end{center}
\end{table*}

\begin{table*}[t!]
\captionsetup{labelsep=newline, font=footnotesize}
\caption{\textsc{Improvement results of CSG\textsuperscript{2}L with three baseline GNN models.}}
\begin{center}
\begin{tabular}{l c c c c c c c}
\toprule
\textbf{Method}&\textbf{Wisconsin}&\textbf{Texas}&\textbf{Cornell}&\textbf{Chameleon}&\textbf{Cora}&\textbf{Citeseer}&\textbf{Average} \\
\midrule
GCN	&57.84±5.69	&59.69±5.18	&50.56±3.09&	58.65±2.73	&85.10±0.62&	73.21±1.26&	64.18 \\
\textbf{CS$\mathrm{\textbf{G}}^2$$\textbf{L}_{\textit{GCN}}$} &	61.37±3.35&	62.16±3.84&	54.28±2.81&	66.19±1.56&	85.59±0.53&	73.73±1.51&	67.22 \\
\textit{improvements}&	↑3.53&	↑2.47	&↑3.72&	↑7.54&	↑0.49	&↑0.52&	↑3.04 \\
\midrule
GIN&	76.35±1.30&	73.87±2.86	&75.40±5.19	&58.53±2.24&	81.17±0.36&	74.35±0.25&	73.28 \\
\textbf{CS$\mathrm{\textbf{G}}^2$$\textbf{L}_{\textit{GIN}}$}&	83.66±2.71	&78.38±1.89	&77.66±3.83	&66.16±1.30	&82.29±0.64	&74.84±0.36	&77.17 \\
\textit{improvements}&	↑7.31	&↑4.51	&↑2.26	&↑7.63&	↑1.12	&↑0.49	&↑3.89 \\
\midrule
GPRGNN & 83.25±0.63 & 81.31±1.03 & 79.93±0.92 & 67.07±0.53 & 85.32±0.26 & 76.06±0.53 & 78.82 \\
\textbf{CS$\mathrm{\textbf{G}}^2$$\textbf{L}_{\textit{GPRGNN}}$} & 86.08±1.53 & 85.57±0.66 & 82.98±0.64 & 68.12±1.02 & 86.89±0.49 & 76.79±0.27 & 81.07 \\
\textit{improvements} & ↑2.83 & ↑4.26 & ↑3.05 & ↑1.05 & ↑1.57 & ↑0.73 & ↑2.25 \\
\bottomrule

\end{tabular}
\label{tab3}
\end{center}
\end{table*}

We compare the proposed model with ten baseline models, including four GNN models: GCN \cite{b5}, GAT \cite{b7}, GIN \cite{b25}, GraphSAGE \cite{b6}, and six recent state-of-the-art special structure GNN models tackling heterophily: MixHop \cite{b27}, H2GCN \cite{b28}, GPRGNN \cite{b26}, GOAL \cite{b29} OrderedGNN \cite{b30}, ES-GNN \cite{b31}. In order to demonstrate the effectiveness and generalizability of our framework, we select three models, GCN, GIN, and GPRGNN, as the enhanced models.

\subsection{Hyper-parameter Setting}
We uniformly set the hyper-parameters of all models as follows: the number of GNN layers is 2, the hidden layer dimension is 64, the temperature parameter is 0.5, the rank of SVD is 5, the dropout rate is set to 0.5, the learning rate is set to 0.05, the weight decay is selected from the set $\{1\mathrm{e}-3, 5\mathrm{e}-3, 5\mathrm{e}-4, 5\mathrm{e}-5, 5\mathrm{e}-6\}$, and the Adam optimizer is adopted. For other hyper-parameters, we use their best default parameters in the original paper. For fair comparison, in the enhanced GNNs(GCN, GIN, GPRGNN), the hyper-parameter settings are the same as those in their counterparts, ensuring a consistent evaluation. We set the weight of contrastive learning loss $\lambda$ in equation \eqref{eq9} to 0.1 and discuss it later.

\subsection{Results on Node Classification}
We use the enhanced model of GPRGNN to perform experimental results on node classification tasks on 6 datasets, and compare with the corresponding baseline and GNN-based SOTA models, as shown in Table~\ref{tab2}. The average accuracy and standard deviation of ten different data splits are presented, and the best results are highlighted in bold. CSG\textsuperscript{2}L achieves significant performance by combining with the SVD-aug module and the LGDL module. It can be observed that although GPRGNN performs well on some datasets (Wisconsin, Texas, Cornell), it is still inferior to the SOTA model on other datasets. Our CSG\textsuperscript{2}L can help GPRGNN obtain the most competitive results, with the best results on four heterophilic datasets, and even a good improvement on homophilic datasets, with an overall performance improvement of 2.25\% on average. This is attributed to the fact that our model can fully combine with local and global information, and further improve the performance by considering the weights of hard and easy sample pairs.

In addition, we also report the improvement results of CSG\textsuperscript{2}L in Table~\ref{tab3}. The results in the table show that CSG\textsuperscript{2}L consistently improves the performance of three baseline GNN models on six datasets. Specifically, the proposed framework improves the performance of GCN by 3.04\% on average, GIN by 3.89\% on average, GPRGNN by 2.25\% on average. This result means that CSG\textsuperscript{2}L is an effective general framework that can improve the performance of well-known GNN models that are already recognized to be effective.

\subsection{Hyper-parameter Analysis}

\begin{figure*}[t!]
\centering
\subfigure
{
    \begin{minipage}[b]{.31\linewidth}
        \centering
        \includegraphics[width=5.63cm, height=4.31cm]{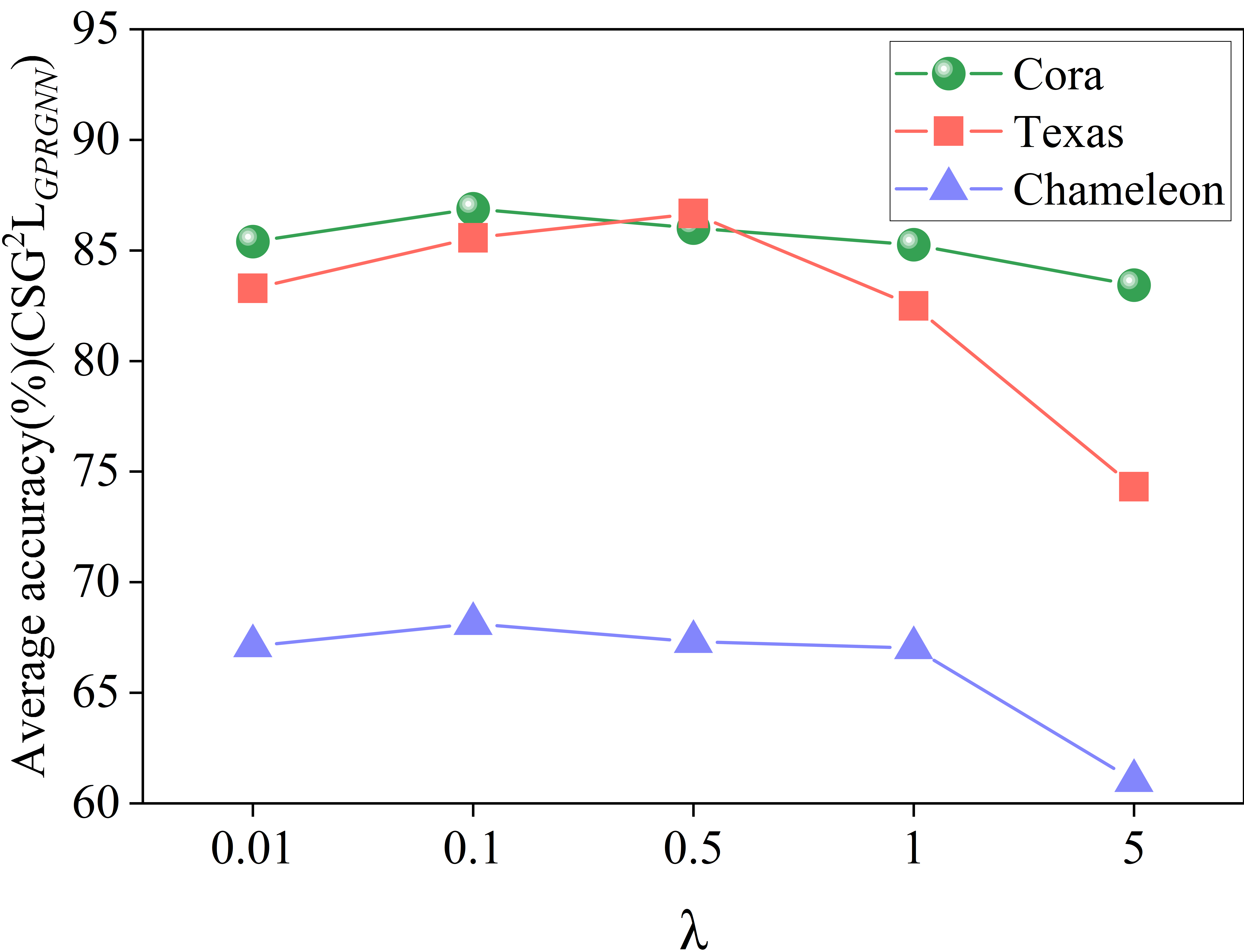}
    \end{minipage}
}
\subfigure
{
 	\begin{minipage}[b]{.31\linewidth}
        \centering
        \includegraphics[width=5.63cm, height=4.31cm]{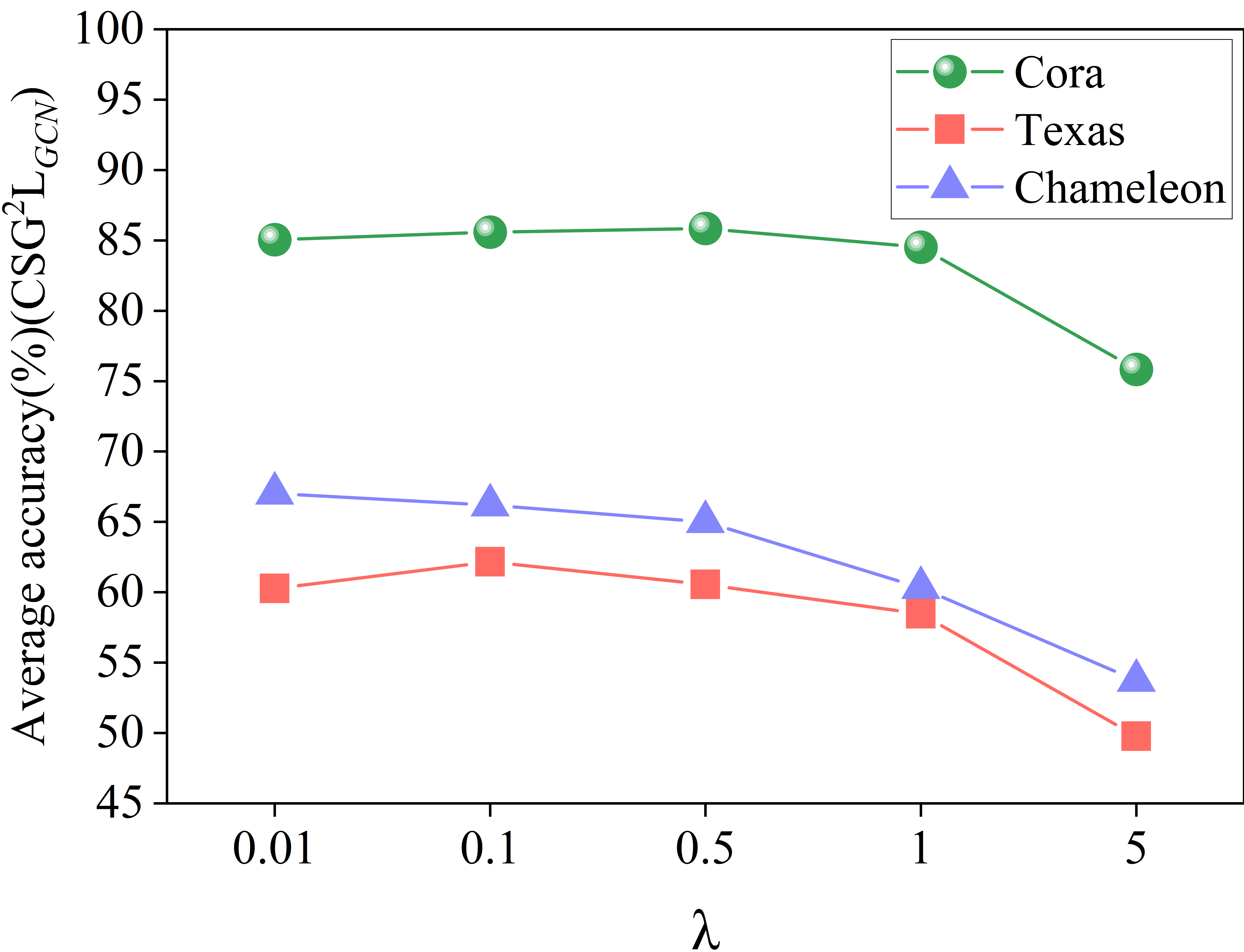}
    \end{minipage}
}
\subfigure
{
 	\begin{minipage}[b]{.31\linewidth}
        \centering
        \includegraphics[width=5.63cm, height=4.31cm]{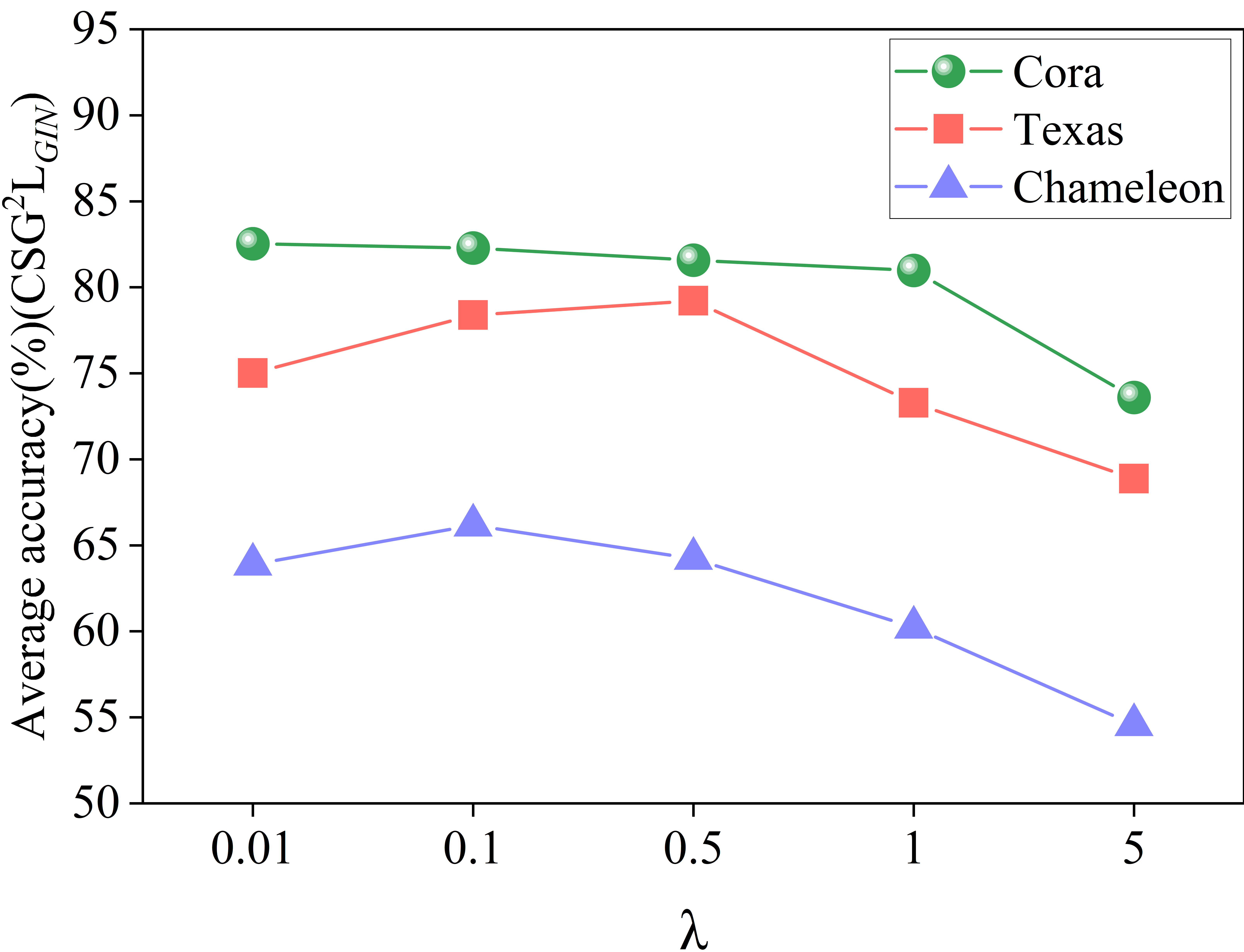}
    \end{minipage}
}
\captionsetup{font=footnotesize}
\caption{Hyper-parameter analysis results of CS$\mathrm{G}^2$Ls on three benchmark datasets.}
\label{fig2}
\end{figure*}

\begin{figure*}[t!]
\centering
\subfigure
{
    \begin{minipage}[b]{.31\linewidth}
        \centering
        \includegraphics[width=5.63cm, height=4.31cm]{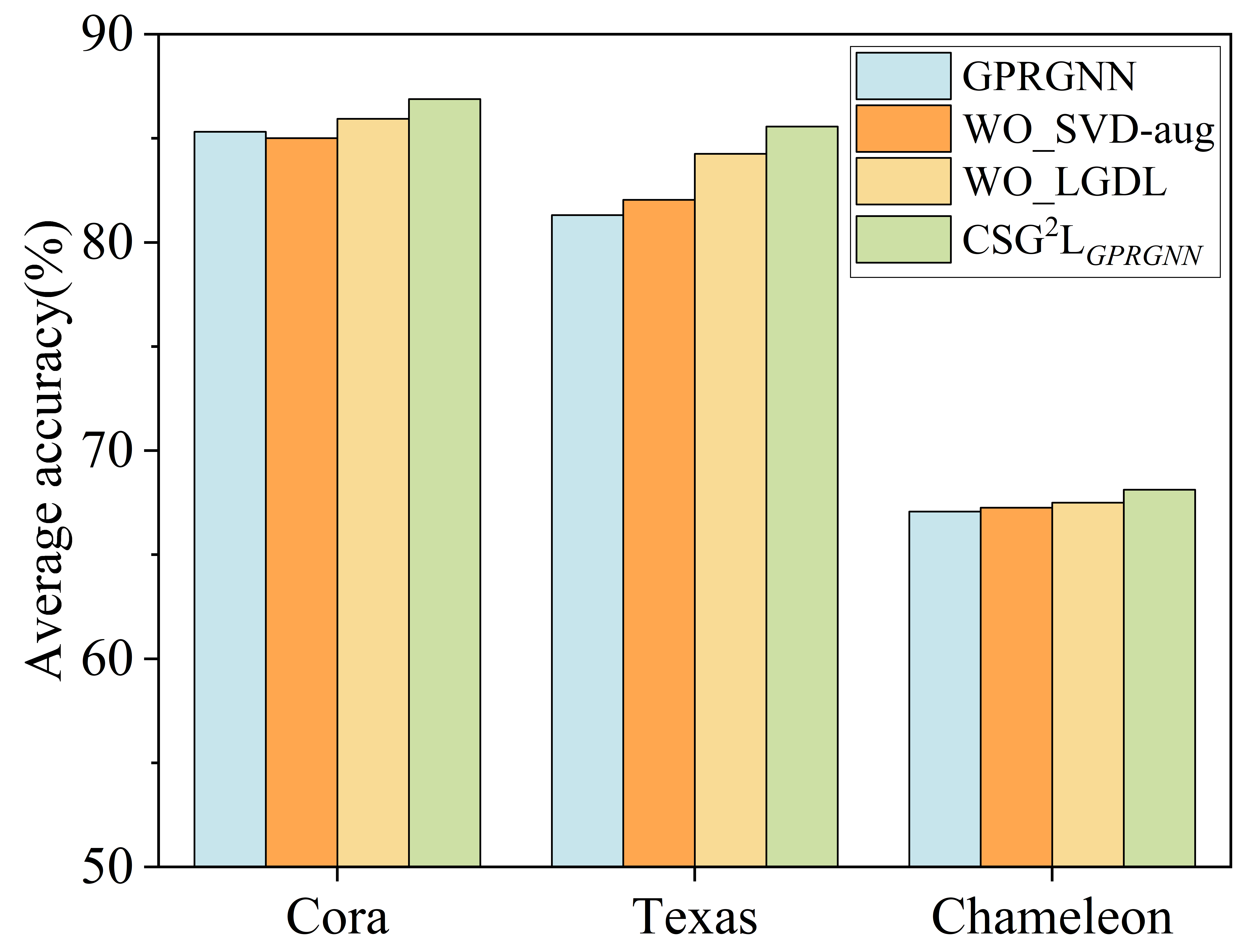}
    \end{minipage}
}
\subfigure
{
 	\begin{minipage}[b]{.31\linewidth}
        \centering
        \includegraphics[width=5.63cm, height=4.31cm]{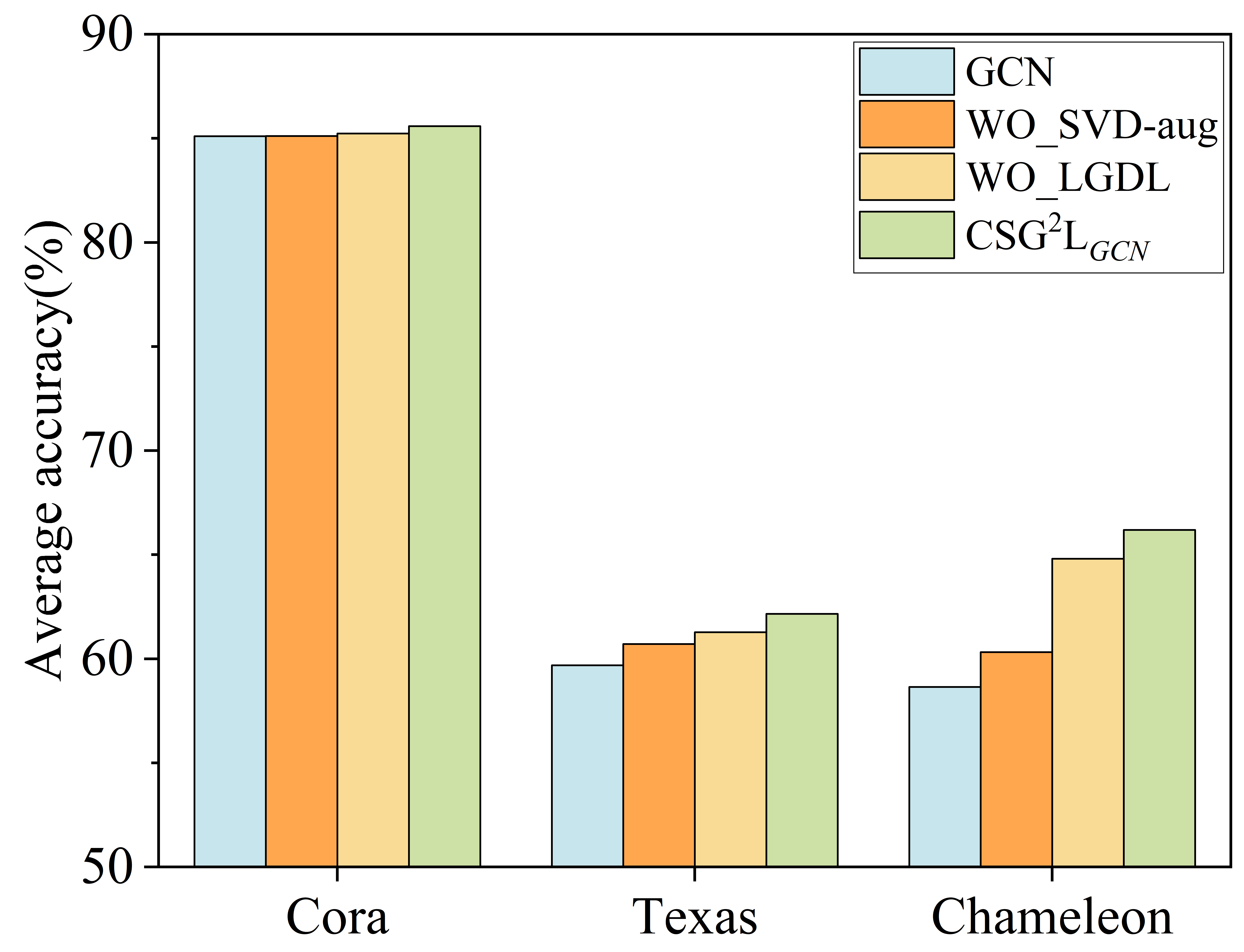}
    \end{minipage}
}
\subfigure
{
 	\begin{minipage}[b]{.31\linewidth}
        \centering
        \includegraphics[width=5.63cm, height=4.31cm]{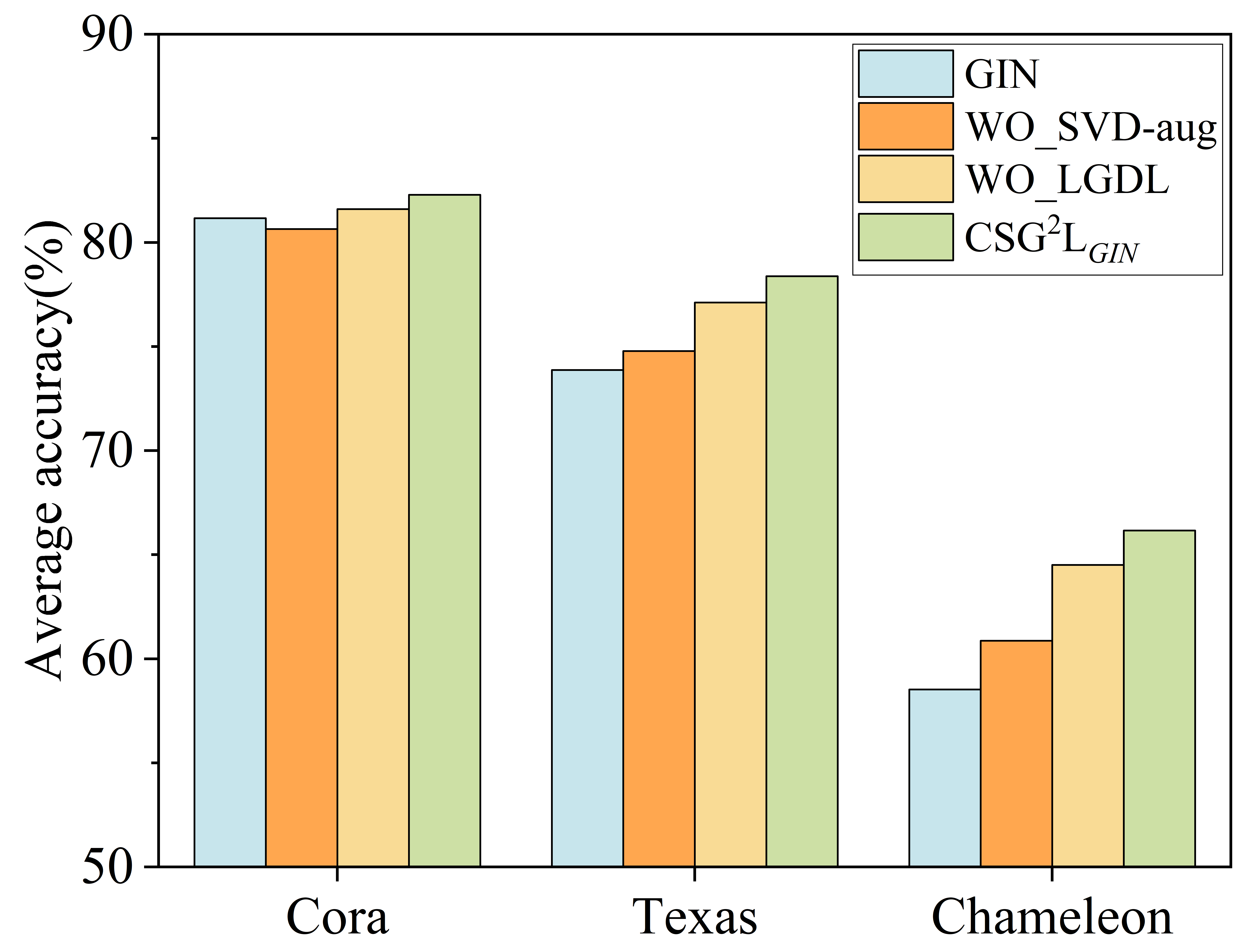}
    \end{minipage}
}
\captionsetup{font=footnotesize}
\caption{Ablation study results on three benchmark datasets.}
\label{fig3}
\end{figure*}
We achieve the simultaneous improvement of the node representation learning quality and classification task performance by jointly optimizing the classification loss and contrastive loss. As defined in equation \eqref{eq9}, the hyper-parameter $\lambda$ is used to adjust the weight of the contrastive loss in the overall loss function, thereby balancing the contribution of the classification task and contrastive learning. In order to deeply analyze the impact of $\lambda$ on model performance, we conduct sufficient experiments using CSG\textsuperscript{2}Ls on three representative benchmark datasets: Cora, Texas, and Chameleon.

The experimental results are shown in Fig.~\ref{fig2}. It can be observed from the figure that when $\lambda$ = 0.01, the performance of our model has not yet reached the optimal level. This is because the weight of the contrastive loss is low at this time, preventing our model from fully utilizing the representation information learned by contrastive learning during the optimization process. As $\lambda$ gradually increases, the model performance gradually improves. When $\lambda$ = 0.1, the model achieves the best average accuracy on the three datasets, indicating that the contrastive loss and classification loss have reached a good balance at this point. In this case, contrastive learning can effectively explore the structural information and global interactions between nodes, prompting the model to learn a more robust node representation without disrupting the optimization of the classification task. However, when $\lambda$ continues to increase, the performance begins to decline significantly, especially when $\lambda$ = 5, the performance drops to the lowest level. This is because the contrastive loss dominates  the overall loss function, resulting in a significant weakening of the contribution of the classification loss. In this case, the model training focuses excessively on bringing the representation of positive sample pairs closer and pulling the representation of negative sample pairs farther away, but ignores the classification objective, which ultimately affects the overall performance. Based on the above analysis, we choose the default value of $\lambda$ = 0.1.

\subsection{Ablation Study}

In order to evaluate the effectiveness of our proposed CSG\textsuperscript{2}L framework in improving the performance of GNN models, we conduct ablation studies on three benchmark datasets. The goal is to analysis the trend of model performance by gradually removing the two different components, and further explore the contribution and importance of each module in the overall framework. The experiment results are shown in Fig.~\ref{fig3}, which shows the ablation study results of the three basic models of GPRGNN, GCN and GIN on three datasets. From the figure, our analysis is as follows:

1) \textit{Original model performance}: In the baseline model, without both SVD-aug and LGDL modules, the model only uses the node feature matrix $X$ and the normalized adjacency matrix $\Tilde{A}$ as inputs for node classification, and the performance is relatively low. This is because the original model does not fully capture global interactions and is easily affected by data sparsity and noise, resulting in poor distinguishability of node representation.

2) \textit{W/O SVD-aug}: In this step, SVD-aug is removed. We introduce random augmentation such as edge perturbations and attribute masking to generate augmented graph, and InfoNCE loss is used to contrast embeddings of the original and augmented graphs, the average performance of the three models is improved. This shows that through simple random perturbations, the model can capture more local structural change information and enhance the robustness of node representation. However, since the transformation introduced by random augmentations lacks pertinence and fails to effectively capture the global collaborative signals, the performance improvement is limited.

3) \textit{W/O LGDL}: In this step, LGDL module is removed. When only introducing the SVD-aug module, the model uses both the original normalized adjacency matrix $\Tilde{A}$ and the reconstructed normalized adjacency matrix $\hat{A}_{SVD}$, along with the node feature matrix $X$ as inputs, and then InfoNCE loss is used to contrast embeddings of the original and augmented graphs. The model performance is further improved because through using low-rank reconstruction to obtain an augmented graph, SVD can effectively capture the global interactions, which makes up for the limitations of random augmentations. Combined with contrastive learning, the embeddings of positive pairs are brought closer, while negative pairs are pushed apart, leading to more discriminative node representations.

4) \textit{CSG\textsuperscript{2}Ls}: On the basis of the SVD-aug module, after introducing the LGDL module with an adaptive reweighting strategy, the three baseline models achieve the best performance. By adaptively adjusting the weights of hard and easy sample pairs, the strategy improves the optimization process of InfoNCE loss and mitigates the effects of noisy and inaccurate contrastive signals. In addition, combined with the SVD-aug module and the LGDL module, CSG\textsuperscript{2}L improves the robustness and effectiveness of representations for classification while maintaining global information.

In summary, the ablation study results demonstrate the effectiveness and necessity of each component in CSG\textsuperscript{2}L. By utilizing the SVD-aug module and the LGDL module, CSG\textsuperscript{2}L significantly improves the performance of the basic GNN models, exhibiting strong generalization ability and robustness in node classification tasks.

\section{Conclusion}
In this paper, we proposed a novel graph contrastive signal generative framework, CSG\textsuperscript{2}L, designed to address the limitations of traditional graph learning methods in node classification tasks. CSG\textsuperscript{2}L contains two components, we build a SVD-aug module which uses low-rank SVD to generate augmented graph that captures global interactions. In addition, we design a LGDL module, which integrates local and global information, optimizing the contrastive learning loss by introducing an adaptive reweighting strategy to further improve the effect of contrastive learning, effectively mitigating the issues of insufficient global interaction extraction and inaccurate contrastive signals. In future work, we plan to integrate attribute information into SVD-aug module to further improve the accuracy of the augmented graph.

\vspace{12pt}

\end{document}